\def\year{2021}\relax
\documentclass[letterpaper]{article} 
\usepackage{aaai21}  
\usepackage{times}  
\usepackage{helvet} 
\usepackage{courier}  
\usepackage[hyphens]{url}  
\usepackage{graphicx} 
\usepackage{natbib}  
\usepackage{caption} 
\usepackage{amsmath}
\usepackage{algorithm}
\usepackage{algpseudocode}
\usepackage{booktabs}
\usepackage{latexsym}
\usepackage{listings}
\usepackage{multicol}
\usepackage{subcaption}
\usepackage{tabularx}
\usepackage{todonotes}

\algnewcommand\algorithmicforeach{\textbf{for each}}
\algdef{S}[FOR]{ForEach}[1]{\algorithmicforeach\ #1\ \algorithmicdo}
\MakeRobust{\Call}

\newcommand{\gammafunc}[1]{\Gamma\left(#1\right)}
\newcommand{\bigO}[1]{O\left(#1\right)}

\newcommand{\latticepoint}{\mathit{latticePoint}}
\newcommand{\relationshiplattice}{\mathit{relationshipLattice}}
\newcommand{\ctt}[1]{\mathit{ct}_{#1}}
\newcommand{\family}{\mathit{family}}
\newcommand{\structurelearning}{\mathit{structureLearning}}
\newcommand{\score}{\mathit{score}}

\newcommand{\C}{\mathit{C}} 
\newcommand{\T}{\mathit{T}} 
\newcommand{\V}{\mathit{V}} 
\newcommand{\K}{\mathit{k}} 

\newcommand{\precount}{PRECOUNT}
\newcommand{\ondemand}{ONDEMAND}
\newcommand{\hybrid}{HYBRID}
\newcommand{\FB}{\textsc{FactorBase}}
\newcommand{\MJ}{M\"obius Join}
\newcommand{\MJFunc}{M\"obiusJoin}
\newcommand{\laj}{learn-and-join}
\newcommand{\contingencytable}{ct-table}
\newcommand{\contingencytables}{ct-tables}
\newcommand{\ct}[2][]{\mathit{ct}_{#1}(#2)}
\newcommand{\defterm}[1]{\textbf{#1}}

\newcommand{\algorithmref}[1]{Algorithm~\ref{#1}}
\newcommand{\equationref}[1]{Equation~\ref{#1}}
\newcommand{\figureref}[1]{Figure~\ref{#1}}
\newcommand{\tableref}[1]{Table~\ref{#1}}

\newcommand{\bayesnet}{BN}
\newcommand{\bayesnets}{BNs}

\newcommand{\sota}{SOTA}

\title{Pre and Post Counting for Scalable Statistical-Relational Model Discovery}
\author {
    Richard Mar,
    Oliver Schulte\thanks{Supported by a Discovery Grant from the NSERC of Canada.} \\
}

\affiliations {
    Simon Fraser University \\
    richard\_mar@sfu.ca, 	oschulte@cs.sfu.ca
}
\begin{document}

\maketitle

\begin{abstract}
Statistical-Relational Model Discovery aims to find statistically relevant patterns in relational data.
For example, a relational dependency pattern may stipulate that a user's gender is associated with the gender of their friends.
As with propositional (non-relational) graphical models, the major scalability bottleneck for model discovery is computing instantiation counts: the number of times a relational pattern is instantiated in a database.
Previous work on propositional learning utilized pre-counting or post-counting to solve this task.
This paper takes a detailed look at the memory and speed trade-offs between pre-counting and post-counting strategies for relational learning. 
A pre-counting approach computes and caches instantiation counts for a large set of relational patterns {\em before} model search.
A post-counting approach computes an instantiation count dynamically on-demand for each candidate pattern generated during the model search. 
We describe a novel hybrid approach, tailored to relational data, that achieves a sweet spot with pre-counting for patterns involving positive relationships (e.g. pairs of users who are friends) and post-counting for patterns involving negative relationships (e.g. pairs of users who are not friends).
Our hybrid approach scales model discovery to millions of data facts.
\end{abstract}

\section{Introduction}

Statistical-Relational Learning (SRL) aims to construct statistical models that extract knowledge from complex relational and network data.
SRL model construction searches through a space of possible models, scoring different candidate models to find a (local) statistical score optimum.
SRL models are typically composed of local dependency patterns, represented with edges in a model graph or clauses in logical syntax.
The main computational burden in model scoring is {\em instance counting}, determining the number of times that a local pattern defined by the model occurs in the dataset.
A key technique to scale counting for large and complex datasets is {\em caching} local instance counts and statistical scores.
Caching trades memory for model search time and is standard in graphical model search for propositional (i.i.d.) data (e.g. ~\cite{tetrad2008a}).
There are two basic approaches to counts caching~\cite{Lv2012}: Pre-counting builds up a large cache of local pattern counts before the model search phase.
Post-counting computes local instance counts only for patterns generated during model search and then adds them (or the resulting scores) to a cache in case the pattern is revisited later during model search.
This paper compares pre and post counting for relational model search.

\paragraph{Why Relational Counting Is  Hard.}
To understand the pros and cons of different count caching strategies, we briefly review why relational counting problems are fundamentally different from, and harder than, counting in propositional data.
For propositional data, a conjunctive pattern specifies a set of attribute values for singletons of entities; for example, the set of female users over 40.
Counting then requires only {\em filtering}, finding the subset of {\em existing} entities matching the pattern.
Relational data specify linked pairs of entities, like users who are friends (assuming binary relationships), where two new counting problems arise:
\begin{enumerate}
\item Relational patterns may involve $k$-tuples of entities/nodes.
For example, a triangle involves 3 nodes.
A naive approach to counting would require enumerating $n^{k}$ $k$-tuples of entities.
In an SQL table representation of relational data, counting for $k$-tuples of entities involves table JOINs; we therefore term it the {\em JOIN problem}.
The JOIN problem has been extensively studied in the database community as part of SQL query evaluation (COUNT(*) queries).
\item Relational patterns may involve {\em negative relationships}, pairs of entities not listed in the data.
For example, we may wish to find the number of women users who have {\em not} rated a horror movie.
We term this the {\em negation problem}.
The JOIN and negation problems both involve {\em pattern instances that are not subsets of the given data tuples.}
\end{enumerate}

\paragraph{Pre vs. Post Counting: Advantages and Disadvantages.}
The main advantage of pre-counting is that the data is scanned only once to build up a cache.
In particular, the number of table JOINs is minimized, as our experiments show.
Also, the cache can be built incrementally using dynamic programming by computing counts for longer patterns from shorter ones. 
For example, in relational data, counts with a relationship chain length of l can be built up from relationship chains of length 1, 2, ..., l - 1 \cite{Yin2004, DBLP:journals/ijdsa/SchulteQ19}.
The main disadvantage of pre-counting is building a large cache of counts for complex patterns, with only a fraction likely to be required for model selection.
For example, pre-counted dependency patterns may involve 20 predicates, whereas model search rarely considers clauses with more than 5.

The main advantage of post-counting is that instances are counted only for patterns generated by the model search.
The disadvantage is that each pattern evaluation requires a new data access with potentially expensive table JOINs. 

A high-level summary of our empirical findings is that because of the JOIN problem, pure post-counting scales substantially worse than pure pre-counting.
At the same time, the negation problem is challenging for the large patterns built up in pre-counting.
Therefore, we introduce a hybrid method that uses pre-counting for the JOIN problem and post-counting for the negation problem.
The hybrid method leverages the \MJ{} algorithm which computes instance counts for relational patterns that may involve positive and negative relationships~\cite{DBLP:conf/cikm/QianSS14}.
The \MJ{} is an inclusion-exclusion technique that requires no further data accesses to the original data, and hence no further table JOINs.

Our contributions may be summarized as follows:
\begin{itemize}
    \item Investigating the strengths and weaknesses of pre/post count counting strategies for different relational dataset properties.
    \item Describing a new hybrid count caching method, tailored towards relational data, that for many datasets combines the strengths of both pre and post counting. Hybrid counting scales to millions of data facts (database rows).
\end{itemize}

\section{Related Work}

We selectively discuss the most relevant topics from previous SRL work. 

{\em Language Bias.}
We consider first-order relational patterns that only involve types of individuals, not specific individuals (e.g. $\mathit{Friend(X,Y)}$ not $\mathit{Friend(joe,Y)}$).
While this is a common restriction in SRL \cite{Ravkic2015,Schulte2016,Kimmig2014}, it limits the applicability of our results to methods that search for clauses involving individuals, such as Boostr~\cite{Natarajan2012}.
Our experiments investigate learning directed models (first-order Bayesian networks), but our observations about the fundamental JOIN and negation problems apply to other SRL models as well.

{\em Search Space: Predicates vs. Clauses.}
Inspired by graphical models, a tradition in SRL is to search for dependencies between predicates, typically represented by links in a model graph.
For example, an edge $\mathit{intelligence(S)} \leftarrow \mathit{grade(S,C)}$ may represent that the grades of a student predicts their intelligence.
Another tradition, inspired by inductive logic programming, searches for dependencies among predicate values, typically represented by clauses (e.g. $ \mathit{intelligence(S) = high}$ :- $\mathit{grade(S,C) = A}$).
Since the predicate space is smaller than that of predicate values, it supports a more efficient model search~\cite{Kersting2007,Friedman99prm,Khosravi2010} at the expense of expressive power~\cite{Khosravi2012a}.
With regard to instance counting, a predicate dependency groups together a set of clauses (one for each combination of parent-child values).
Model evaluation requires instance counting for the entire clause set, which is often more efficient than separate counting, if the data for the same predicates is stored together.
For example if in an SQL table, grades are stored in the same table column, then counting instances for different grades requires only a single table JOIN.
While our experiments evaluate only predicate-level search, we expect that the general trends apply also to clause-level search.

{\em Approximate vs. Exact Counting.}
Several SRL approaches have increased scalability by aiming for approximate counting, which is often sufficient for model evaluation (e.g. \cite{das2019fast}).
In this paper we use only exact counts, obtained by SQL queries.
Other SRL work has also shown the usefulness of SQL optimization for obtaining instance counts (e.g.~\cite{Niu2011}).
Approximate counting speeds up the instance for a {\em single pattern}, which is orthogonal to our topic of counts caching; caches can be used for both approximate and exact counts.
We discuss approximate counting further in the future work section.
Another type of approximation is negative sampling of unlinked node pairs, which adds extra node pairs to the data~\cite{Nickel2016}.
Computationally, negative sampling solves the negation problems but increases the JOIN problem.
Caching counts for negative sampling is an interesting direction for future work.


\section{Graphical Model Background}
\begin{figure}[t]
    \centering
    \includegraphics[width=0.90\columnwidth]{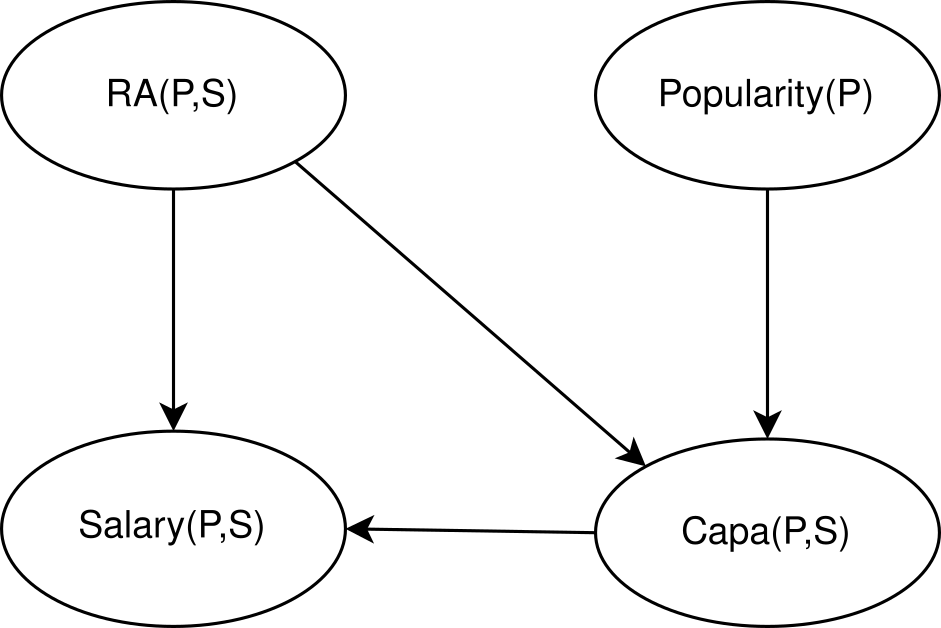}
    \caption{An example of a first-order Bayesian network.}
    \label{figure:first-order-bayesnet}
\end{figure}

We choose the \FB{} system \cite{DBLP:journals/ijdsa/SchulteQ19} for experimentation and to illustrate computational counting challenges.
\FB{} learns a first-order Bayesian network (\bayesnet{})~\cite{Poole2003,Wang2008} for a relational dataset.
An example of a first-order Bayesian network can be found in \figureref{figure:first-order-bayesnet}.
\FB{} is an appropriate benchmark as it achieves \sota{} scalability~\cite{Schulte2016}.
We focus on our description on fundamental issues associated with relational counting that we expect to generalize to other systems.
In the Limitations section, we discuss design options not explored in our experiments.

As an example model score, the BDeu score for a given \bayesnet{} $B$ and dataset $D$ is defined in \equationref{eq:bdeu}.
\begin{figure*}[t]
\begin{equation}
    BDeu(B, D) =
    \log{\left(P(B)\right)}
    +
    \sum_{i=1}^{n}
    \sum_{j=1}^{q_i}
    \left(
        \log
        {\left(
            \frac
            {
                \gammafunc
                {
                    \frac
                    {
                        N'
                    }
                    {
                        q_i
                    }
                }
            }
            {
                \gammafunc
                {
                    N_{ij}
                    +
                    \frac
                    {
                        N'
                    }
                    {
                        q_i
                    }
                }
            }
        \right)}
        +
        \sum_{k=1}^{r_i}
        \log
        {\left(
            \frac
            {
                N_{ijk}
                +
                \frac
                {
                    N'
                }
                {
                    r_{i}q_i
                }
            }
            {
                \gammafunc
                {
                    \frac
                    {
                        N'
                    }
                    {
                        r_{i}q_i
                    }
                }
            }
        \right)}
    \right)
    \label{eq:bdeu}
\end{equation}
\end{figure*}
A description of the terms can be found in \tableref{table:variables:bdeu}.
A \defterm{parent configuration} for node $i$ is a tuple of values, one for each parent.

\paragraph{Example.}
Let node $i$ be $\mathit{Salary(P,S)}$ with 4 possible values $r_i$ (LOW, MED, HIGH, N/A).
Together with its parents, node $i$ forms the model {\em family}, which expresses a relational dependency pattern $\mathit{RA(P,S), Capa(P,S)} \rightarrow$ \\ $\mathit{Salary(P,S)}$.
Suppose we assign index $j$ to the parent configuration $\mathit{RA(P,S) = T, Capa(P,S) = 4}$.
Then $N_{ij}$ counts how often this configuration occurs in the dataset $D$: the number of professor-student pairs such that the student is an RA for the professor with a high capability of 4.
Suppose we assign index $k$ to the salary value HIGH.
Then $N_{ijk}$ counts how often the parents are in configuration $j$ and the child node takes on value $k$: the number of professor-student pairs such that the student is an RA for the professor with a high capability of 4 and receives a HIGH salary. 
In \tableref{table:ct}, this count equals 5.
In terms of clauses, $N_{ijk}$ represents the instantiation count of the clause $\mathit{RA(P,S) = T}, \mathit{Capa(P,S) = 4 \rightarrow Salary(P,S) = HI}$-$\mathit{GH}$.

\begin{table}[t]
\centering
\begin{tabularx}{\columnwidth}{@{}rX@{}}
\toprule
Variable   & Description                                   \\
\midrule
$n$        & Number of nodes in the model graph           \\
$r_i$      & Number of possible values for node $i$       \\
$q_i$      & Number of parent configurations for node $i$ \\
$N^\prime$ & Equivalent sample size.                       \\
$N_{ij}$   & Number of occurrences where node $i$ has its 
             parents in the $j^{th}$ configuration.        \\
$N_{ijk}$  & Number of occurrences where node $i$ is
             assigned its $k^{th}$ value and its parents are assigned the $j^{th}$ configuration.                \\
\bottomrule
\end{tabularx}
\caption{Terms in the BDeu scoring metric, which is used to score \bayesnets{} such as the one found in \figureref{figure:first-order-bayesnet}.
}
\label{table:variables:bdeu}
\end{table}
\citeauthor{Schulte2017a} showed how the BDeu score can be adapted for multi-relational data \citeyearpar{Schulte2017a}.
The count operations required are essentially the same between the relational BDeu score and the propositional score shown, and similar for other \bayesnet{} scores.
The most computationally expensive part of BDeu, and many other scoring metrics, is generating instantiation or frequency counts to obtain the $N_{ij}$ and $N_{ijk}$ values.
Instance counts can be represented in a \textbf{contingency table (\contingencytable{})}: for a list of variables $V_1,\ldots,V_m$, the table contains a row for each value tuple $v_1,\ldots,v_m$, and records how many times this value combination occurs in the data set~\cite{Lv2012,DBLP:conf/cikm/QianSS14}.
Following~\cite{Lv2012}, {\em we use \contingencytables{} as caches for instance counts.} 

\begin{figure}[t]
    \centering
    \vspace{0.5em}
    \includegraphics[width=\columnwidth]{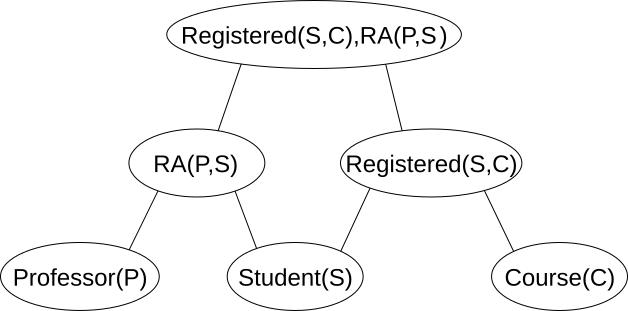}
    \caption{A relationship lattice for two relationships (students register in courses and work as RAs for professors) where each node in the graph is a lattice point.}
    \label{figure:relationship-lattice}
\end{figure}

\begin{table}[t]
    \centering
    \begin{tabular}{lll}
        \toprule
        Positive \contingencytable{} & Negative \contingencytable{}         & Learning Method \\
        \midrule
        Lattice Point   & Lattice Point & \precount{} \\
        Family          & Family        & \ondemand{} \\
        Lattice Point   & Family        & \hybrid{} \\
        Family          & Lattice Point & IMPOSSIBLE \\
        \bottomrule
    \end{tabular}
    \caption{Summary of possible methods to consider for computing \contingencytables{} from a relational
database based on the inputs for creating the positive and negative \contingencytables{}.}
    \label{table:methods}
\end{table}

\section{Computing Relational Contingency Tables}

We use the method of \cite{DBLP:conf/cikm/QianSS14}, as implemented in \FB{}.
As discussed under Related Work, other methods may be employed as well, such as approximate counting.
In a relational \contingencytable{}, a key role is played by relationship indicator variables that indicate whether a relationship holds (e.g. $\mathit{Registered(S,C)} = T$ or $\mathit{Registered(S,C)}$) or not (e.g. $\mathit{Registered(S,C)} = F$ or $\neg \mathit{Registered(S,C)}$).
In a \textbf{positive} \contingencytable{}, all relationship indicator variables are set to $T$ (true).
A complete \contingencytable{} contains values for both true and false relationships; see \tableref{table:ct} for illustration.
The last 9 rows form the positive  \contingencytable{}.
The predicate \emph{RA(P,S)} represents that a student is an RA for a professor, and the attributes \emph{Capa(P,S)} and \emph{Salary(P,S)} represent a student's salary and capability when they are the RA for a given professor.

\begin{table}[t]
    \centering
    \begin{tabular}{cccc}
    \toprule
    \textbf{Count} & \emph{Capa(P,S)} & \emph{RA(P,S)} & \emph{Salary(P,S) } \\
    \midrule
    203   & N/A   & F     & N/A \\
    5     & 4     & T     & HIGH \\
    4     & 5     & T     & HIGH \\
    2     & 3     & T     & HIGH \\
    1     & 3     & T     & LOW \\
    2     & 2     & T     & LOW \\
    2     & 1     & T     & LOW \\
    2     & 2     & T     & MED \\
    4     & 3     & T     & MED \\
    3     & 1     & T     & MED \\
    \bottomrule
    \end{tabular}%
    \caption{A contingency table represents instantiation counts of a conjunctive condition in a database. For example, the last row shows that the number of instances (groundings) is 3 for the assignment \emph{Capa(P,S)} = 1, \emph{RA(P,S)} = T, \emph{Salary(P,S)} = MED. Numbers are chosen for illustration.}
    \label{table:ct}%
\end{table}%

\FB{} uses a 2-stage approach to compute a relational \contingencytable{}, which is described as follows:

\begin{enumerate}
\item Input: A list of variables $V_1,\ldots,V_m$ and a relational dataset $D$.
    \item Generate a positive \contingencytable{} $\ct[+]{\mathbf{R}}$ for each relationship lattice point $\mathbf{R}$. \FB{} uses SQL INNER JOINs to compute a \contingencytable{} for existing relationships (a COUNT(*) query with a GROUP BY clause). 
    \item Extend  $\ct[+]{\mathbf{R}}$ to a complete \contingencytable{} $\ct{\mathbf{R}}$. \FB{} uses the \MJ{} for this step.
\end{enumerate}

The \MJ{} is an inclusion-exclusion technique whose details 
are somewhat complex and not necessary for this work.
The main points of importance for this research are as follows:

\begin{enumerate}
    \item Given a positive \contingencytable{}, where all relationships are true, the \MJ{} returns a \contingencytable{} for both existing and non-existing relationships {\em without further access of the original data.}
    \item The runtime cost of the \MJ{} as shown by \cite{DBLP:conf/cikm/QianSS14} is the following:
\begin{equation}
    \bigO{r\log{\left(r\right)}}
    \label{eq:moebius-join:bigO}
\end{equation}
    where $r$ is the number of rows in the output \contingencytable{}. 
\end{enumerate}
%
Next we consider approaches for computing relational \contingencytables{}.
The basic choices are to compute a {\em global} \contingencytable{} for all variables.
Or a smaller \contingencytable{} for those associated with a family (local pattern).
\tableref{table:methods} lays out the options, which we discuss in the next sections.

\section{\precount{} Counts Caching Method}
\label{section:precount-method}

Chains of relationships form a lattice that can be used to structure relational model search~\cite{DBLP:journals/ml/SchulteK12,Yin2004,Friedman99prm}; see ~\figureref{figure:relationship-lattice}.
For instance, the \laj{} model search~\cite{DBLP:journals/ml/SchulteK12} builds up \bayesnets{} for each lattice point in a bottom-up fashion.
The pre-count method computes a \contingencytable{} for each relationship chain, for all variables associated with the chain.
For example, for the lattice point $\mathit{Registered(S,C)}$, the variables include all attributes of students, courses, and the registration link (e.g. $\mathit{grade})$.
The high-level structure of pre-count caching is illustrated in \algorithmref{alg:precount}.
Given a large \contingencytable{}, we can compute a smaller \contingencytable{} for a subset of columns by summing out the unwanted columns; this operation is called \textbf{projection}~\cite{Lv2012}.
For example, given a \contingencytable{} for the lattice point $\mathit{Registered(S,C)}$ with columns for all attributes of students, projection can be used to obtain a \contingencytable{} for just one student attribute.
During structure search, pre-count computes \contingencytables{} for local families from the applicable lattice point \contingencytable{} using projection.

\precount{} will generate  \contingencytables{} for each relationship chain that grow in size as we increase the chain length, which is a key factor for its computational performance.

\begin{algorithm}
    \begin{algorithmic}[1]
        \ForEach{$\latticepoint \; LP \in \relationshiplattice$}
            \State $\ctt{+}(LP) \gets \Call{InnerJoin}{\Call{Tables}{LP}}$
            \State $\ctt{}(LP) \gets \Call{\MJFunc}{ \ctt{+}(LP)}$
        \EndFor
        \ForEach{$\family \in \structurelearning$}
         \State $\ct{\family} \gets \Call{Project}{\ct{LP}, \family}$
            \State $\score \gets \Call{BDeu}{\ct{\family}}$
        \EndFor
    \end{algorithmic}
    \caption{The \precount{} method: pre-compute \contingencytables{} 
    for each lattice point.}
    \label{alg:precount}
\end{algorithm}

\subsection{Contingency Table Growth Rate for \precount{}}

Given a database table $\T$ with $\C$ columns where each column contains $\leq \V$ values each, the size (number of rows) of $\T$ can be upper bound using the following expression:
\begin{equation}
    \Call{Size}{T} = \bigO{V^C}
    \label{eq:tablesize:precount}
\end{equation}
From \equationref{eq:tablesize:precount} we see that \contingencytables{} grow exponentially with respect to the number of columns when using \precount{}.
The alternative is to generate many small tables that are equivalent to the single large one, but grow in size at a slower rate.
This is the approach used by \ondemand{}.

\section{\ondemand{} Counts Caching Method}

This is a relational adaptation of the single-table post-counting concept from \cite{Lv2012}.
\ondemand{} 
computes \contingencytables{} for {\em families} not relationship chains as \precount{} does.
Family \contingencytables{} are much smaller than relationship chain \contingencytables{}, roughly constant-size in practice.
\ondemand{} generates a \contingencytable{} for each family of nodes that is scored.
%
\begin{algorithm}
    \begin{algorithmic}[1]
        \ForEach{$\family \in \structurelearning$}
            \State $\ctt{+}(\family) \gets \Call{InnerJoin}{\Call{Tables}{\family}}$
            \State $\ctt{}(\family) \gets \Call{\MJFunc}{\ctt{+}(\family)}$
            \State $\score \gets \Call{BDeu}{\ct{\family}}$
        \EndFor
    \end{algorithmic}
    \caption{The \ondemand{} method: compute \contingencytables{} for each family during structure search.}
    \label{alg:ondemand}
\end{algorithm}
\subsection{Contingency Table Growth Rate for \ondemand}
\label{subsection:ondemand-ct-table-growth-rate}

Given a table $\T$ with $\C$ columns, with each column containing $\leq \V$ values each and $\K$ parents to choose from, the total size of the equivalent multi-table version of the single large one can be upper bound using the following expression:
\begin{equation}
    \Call{Size}{T} = \bigO{\C \binom{\C - 1}{k} V^{k + 1}} 
    \label{eq:tablesize:postcount}
\end{equation}
From \equationref{eq:tablesize:postcount} we see that {\em \contingencytables{} grow exponentially with respect to the size of the family of nodes being scored when using \ondemand{}.}
Therefore, \ondemand{} is advantageous only if the family sizes generated during search are small.
Several facts support that the family sizes generated will be small.
The BDeu equation (\equationref{eq:bdeu}) shows that large families incur an exponential penalty in the model score; the same holds for other standard scores.
The literature on \bayesnet{} learning states that empirically, the maximum number of parents is typically 4~\cite{Lv2012, Tsamardinos2006}.
For the datasets in our study, we also observe that the \bayesnet{} nodes have small indegrees. 
\tableref{table:databases} shows the mean number of parents per node in the \bayesnet{} learned for the various databases used in the experiments.
While the size of each \contingencytable{} generated by \ondemand{} is small, the method generates many positive \contingencytables{} and as a result executes many expensive table JOINs.

\section{\hybrid{} Counts Caching Method}
\label{section:hybrid-method}

This is our new alternative method for 
computing \contingencytables{} from a relational database.
The pseudocode for it is found in \algorithmref{alg:hybrid}.
\hybrid{} combines the strengths from \precount{} and \ondemand{} and has the following characteristics:
\begin{itemize}
    \item Like \precount{}, it generates a positive \contingencytable{} $\ct[+]{\mathbf{R}}$ for each relationship lattice point $\mathbf{R}$.
    \item Like \ondemand{}, it generates a complete \contingencytable{} $\ct{\mathbf{F}}$ for each family of nodes $\mathbf{F}$ being scored.
\end{itemize}
\begin{algorithm}
    \begin{algorithmic}[1]
        \ForEach{$\latticepoint \; LP  \in \relationshiplattice$}
            \State $\ct[+]{LP} \gets \Call{InnerJoin}{\Call{Tables}{LP}}$
        \EndFor
        \ForEach{$\family \in \structurelearning$}
            \State $\ct[+]{\family} \gets \Call{Project}{\ct[+]{LP}, \family}$
            \State $\ctt{}(family) \gets \Call{\MJFunc}{\ct[+]{\family}}$
            \State $\score \gets \Call{BDeu}{\ctt{}(family)}$
        \EndFor
    \end{algorithmic}
    \caption{The \hybrid{} method: pre-compute positive \contingencytables{} for each lattice point and compute \contingencytables{} for each family during structure search.}
    \label{alg:hybrid}
\end{algorithm}

The \hybrid{} method uses the large relationship chain \contingencytable{} $\ct[+]{\mathbf{R}}$ as a cache to replace expensive JOINs with projections (line 5).
Assuming that local families are small, extending positive \contingencytables{} to complete ones is relatively fast; see \equationref{eq:moebius-join:bigO}.
Both \hybrid{} and \precount{} assume that the overall number of columns/relationships in the database is moderate so that computing the positive \contingencytable{} $\ct[+]{\mathbf{R}}$ using all columns of a given lattice point is feasible.
If the overall number of columns/relationships is too large to compute the $\ct[+]{\mathbf{R}}$ tables, \ondemand{} must be used instead of \precount{} or \hybrid{}.

\section{Empirical Evaluation}

\subsection{Hardware and Datasets}
\label{section:hardware-and-datasets}

Jobs were submitted to Compute Canada's Cedar high performance cluster using Slurm version 20.11.4.
Each job requested one 2.10GHz Intel Broadwell processor and enough RAM to process each dataset.
Using MariaDB version 10.4 Community Edition, 8 real databases of varying size and complexity were supplied as input to \FB{} for each method in \tableref{table:methods}.
Of these 8 databases, 7 are benchmark databases that have been used in previous studies where the scalability of methods for constructing \bayesnets{} from relational databases was studied \cite{DBLP:journals/ijdsa/SchulteQ19}.
Visual Genome is added to this list of benchmark databases due to the large amount of data and number of relationship tables it possesses compared to the other 7 databases; see  \tableref{table:databases}.
Visual Genome is the only dataset that required preprocessing where ternary relationships had to be converted into binary relationships using the standard star schema approach~\cite{Ullman1982}.

\begin{table}
    \centering
    \begin{tabularx}{\columnwidth}{@{}llcc@{}}
    \toprule
    Database & Row Count & \# Relationships & MP/N \\
    \midrule
    UW            & 712      & 2 & 1.6   \\
    Mondial       & 870      & 2 & 1.3   \\
    Hepatitis     & 12,927    & 3 & 1.7  \\
    Mutagenesis   & 14,540    & 2 & 1.6  \\
    MovieLens     & 74,402    & 1 & 1.4  \\
    Financial     & 225,887   & 3 & 1.9  \\
    IMDb          & 1,063,559  & 3 & 3.4 \\
    Visual Genome & 15,833,273 & 8 & 0.5 \\ 
    \bottomrule
    \end{tabularx}
    \caption{Databases used as input to \FB{} with their associated total number of table rows, number of relationship tables they contain, and the mean number of parents per node (MP/N) in the \bayesnets{} learned when using \FB{} and the BDeu score.}
    \label{table:databases}
\end{table}

\begin{table}
\centering
\begin{tabularx}{\columnwidth}{@{}lXX@{}}
\toprule
DB           & Estimated Total $\ct{\mathit{family}}$ Row Count & $\ct{\mathit{database}}$ Total Row Count \\
\midrule
MovieLens     & 816           & 239         \\
Mutagensis    & 6075          & 1631        \\
UW            & 15318         & 2828        \\
Visual Genome & 2923968       & 20447       \\
Mondial       & 55800         & 1738867     \\
Financial     & 930468        & 3013006     \\
Hepatitis     & 176220        & 12374892    \\
IMDb          & 33040         & 15537457    \\
\bottomrule
\end{tabularx}
\caption{Size of the \contingencytables{} generated for each family scored and the entire database.  The $\ct{family}$ column is for \ondemand{} and \hybrid{} while the $\ct{database}$ column is for \precount{}.  Databases are sorted in ascending order based on the size of the \contingencytable{} generated for the entire database.}
\label{table:ct-metrics}
\end{table}

\subsection{Runtime Metrics Reported}

Since the focus of this research is reducing the time it takes to generate \contingencytables{} for large relational databases, we break down \contingencytable{} construction time into the following 3 components: (1) Metadata (2) Positive \contingencytable{} ($\ctt{+}$) (3) Negative \contingencytable{} ($\ctt{-}$).
%
\begin{figure*}[t]
    \centering
    \includegraphics[width=\textwidth]{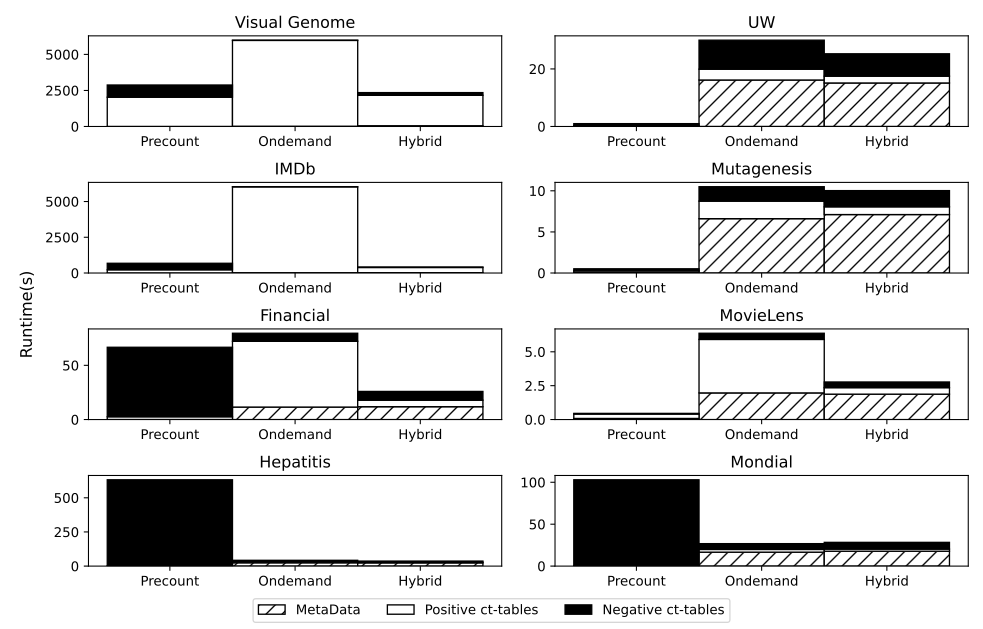}
    \caption{Comparison of the time it takes for \contingencytables{} to be constructed for \precount{}, \ondemand{}, and \hybrid{}.  For IMDb and Visual Genome, \ondemand{} failed to complete within the allotted time so only partial results have been included for those datasets in this figure.}
    \label{figure:methods-summary}
\end{figure*}
\paragraph{MetaData}

This component consists of various metadata specifying the relational schema (syntax) that is extracted and generated from the input database.
It includes the extraction of the first-order logic information related to the \contingencytables{} being generated, generation of the relationship lattice, and generation of the metaqueries, which are used to construct the dynamic SQL query statements for performing the necessary database queries.

\paragraph{Positive \contingencytable{}}

This component consists of the $\ctt{+}$ counting information, which is made of two types of counting queries.
Counts for entity tables are generated with a simple GROUP BY query involving no JOINs.
To generate positive \contingencytables{} for relationship chains, we used a GROUP BY SQL query with INNER JOINs.

\paragraph{Negative \contingencytable{}}

This component consists of the \MJ{} so it is made of the time it takes to generate the counts for all combinations of negative relationships.

\subsection{Runtime Results}

\figureref{figure:methods-summary} shows the cumulative time it takes to construct all the \contingencytables{} for a given database.
The times are broken down into the 3 components of MetaData, Positive \contingencytable{} ($\ctt{+}$), and Negative \contingencytable{} ($\ctt{-}$) as described previously.
For \ondemand{}, the execution of \FB{} exceeded a runtime limit of 100 minutes on IMDb and Visual Genome.
Hence their results are omitted, or are incomplete if included in a figure.
As expected, due to the large size of the \contingencytables{} being generated, \precount{} requires substantially more time 
to generate the negative \contingencytable{} than the two comparison methods.
Overall \ondemand{} performs much worse than \precount{}, mostly due to the time it takes to generate the local positive \contingencytables{}, especially for the large databases like IMDb and Visual Genome.
\hybrid{} on the other hand performs better than \precount{} on several of the databases.
We analyse this improvement in detail next.

\paragraph{Negative \contingencytables{}}

The reason for the general speed-up in negative \contingencytable{} construction time is that \hybrid{} generates smaller \contingencytables{} (see Equations \ref{eq:moebius-join:bigO} and \ref{eq:tablesize:postcount}). 

Some databases are exceptions to this trend (UW, Mutagenesis and MovieLens).
Here the local \contingencytables{} are similar in size to the global one, so the number of \contingencytables{} dominates the relative performance (rather than their size).
For further evidence of this explanation, \tableref{table:ct-metrics} shows that for the databases where \hybrid{} does worse than \precount{}, the total number of rows for the \contingencytables{} generated when using \hybrid{} exceeds the number of rows for the global \contingencytable{} generated when using \precount{}.

\paragraph{Positive \contingencytables{}}

The positive \contingencytables{} for each family scored by \hybrid{} are generated by {\em projecting the information from the full positive \contingencytables{}} for the lattice points.
Therefore, scoring a family does not require further table JOINs. 
Although \hybrid{} performs better than \precount{} for most of the databases experimented with, there are a few cases where \precount{} does better.
This is mainly a result of the MetaData component for \hybrid{} taking substantially longer due to overhead costs, and also due to
the Negative \contingencytable{} component.

\paragraph{MetaData}

\figureref{figure:methods-summary} reveals that \hybrid{} inherited \ondemand{}'s increased overhead to generate the metadata, which is shown to be negligible for \precount{}.
Although the metadata overhead is a bottleneck only for \ondemand{} and \hybrid{}, it does not hurt scalability:
the Metadata component is prominent for small databases like Mondial and UW, but essentially disappears for large databases like IMDb and Visual Genome.

\subsection{Memory Profiling}
\label{section:memory-profiling}

\begin{figure}
    \centering
    \includegraphics[width=\columnwidth]{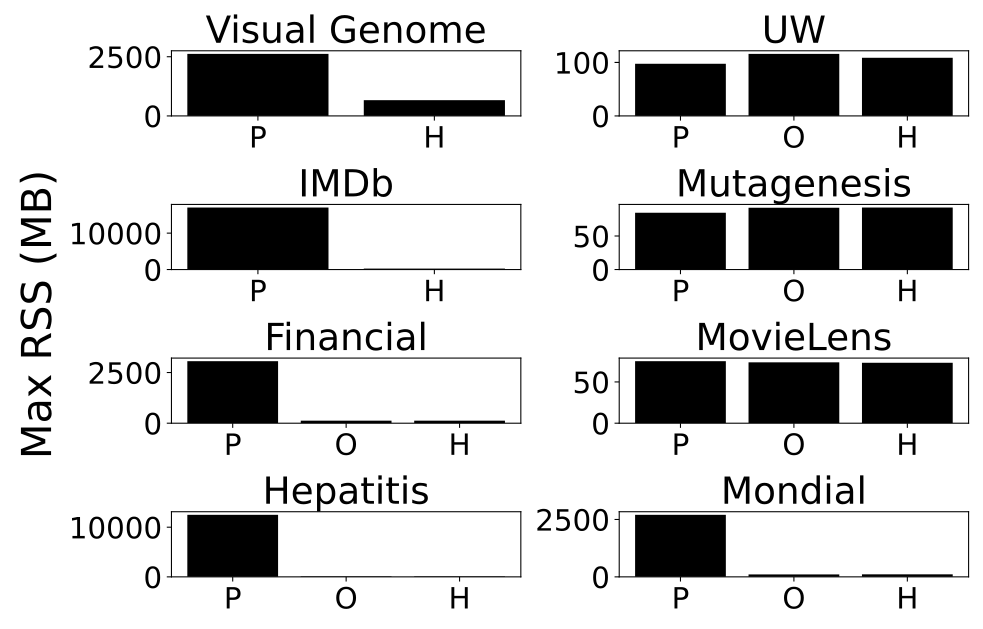}
    \caption{Comparison of the maximum resident set size used by the Java portion of \FB{} when constructing \contingencytables{} for \precount{} (P), \ondemand{} (O), and \hybrid{} (H).  Results have been omitted for \ondemand{} in the cases where it failed to complete within the allotted time.}
    \label{figure:peak-memory}
\end{figure}

\figureref{figure:peak-memory} compares the peak memory used when processing the various databases.
It shows that {\em \precount{} is generally more memory intensive than \ondemand{} and \hybrid{}}.
However, there are 3 databases where \precount{} uses similar amounts of memory.
These 3 databases correspond to the ones in \figureref{figure:methods-summary} where \precount{} performs the best for the Negative \contingencytable{} component.
In addition, \tableref{table:ct-metrics} shows in these 3 databases the total size of the \contingencytables{} generated using \ondemand{} and \hybrid{} exceeded the size of the global \contingencytable{} generated by \precount{} for the entire database.
This observation suggests that the reason why \precount{} does better on these databases is that they are all generating negative \contingencytables{} similar in size.
Recall by \equationref{eq:moebius-join:bigO} the time to generate \contingencytables{} depends on the size of the tables being generated.
Also recall that \ondemand{} and \hybrid{} generate more 
\contingencytables{} than \precount{}, which explains their greater runtime for generating the negative \contingencytables{}.

\section{Limitations and Future Work}

We discuss parts of the count-cache design space that we have not covered in our experiments.

{\em Pre-Count Variants.} Instead of storing instance counts for complete patterns, an intermediate approach is to store additional local information before model search that facilitates post-counting. An example is tuple ID propagation~\cite{Yin2004}.
The general idea of tuple ID propagation is that instead of performing an expensive JOIN operation, for each node and relationship type, we store the set of linked node IDs.
A table JOIN is then replaced by ID propagation.
Compared to the lattice pre-count approach, tuple ID propagation scales well in the number of data columns (predicates) but less well in the number of nodes (rows).
Tuple ID propagation may be suitable for expanding the search space to include clauses with individuals since it expands the data with individual information.

{\em Approximate Counting} may make a purely post-counting method feasible even for large datasets.
Our experiments suggest that it would have to be orders of magnitude more efficient than SQL JOINs 
to make \ondemand{} competitive with \precount{}.

\section{Conclusion}

The key computational burden in learning SRL models from multi-relational data is obtaining the pattern instantiation counts that summarize the sufficient statistics (frequencies of conjunctions of relevant predicate values).
Our research has investigated two approaches to caching relational counts, before and during SRL model search.
Each approach has different implications for two counting challenges fundamental to relational learning: the JOIN problem (counting for chains of existing relationships) 
and the negation problem (counting for non-existing relationships).
We find that for many databases, the most scalable solution is a hybrid approach, which uses pre-counting before model search for the JOIN problem, and post-counting for the negation problem.

\bibliography{mar-paper}
\end{document}